\documentclass[pre,twocolumn,superscriptaddress]{revtex4}
\usepackage{amsmath}
\usepackage{amssymb}

\usepackage{amsmath,amssymb}
\usepackage[usenames]{color}
\usepackage{amssymb}
\usepackage{grffile}
\usepackage{amsmath, amstext, amssymb, amsfonts, amsxtra}
\usepackage{textcomp}
\usepackage{xspace} 
\usepackage[colorlinks]{hyperref}
\usepackage{tikz}
\usetikzlibrary{shapes,arrows,fit,positioning}

\usepackage{cleveref}

\newcommand{\im}{{\rm i}}
\newcommand{\real}{{\rm Re}}
\newcommand{\imag}{{\rm Im}}
\newcommand{\Rdomin}{\mathbb{R}}
\newcommand{\Cdomin}{\mathbb{C}}

\newcommand{\adj}[1]{\tilde{#1}}
\newcommand{\conj}[1]{{#1}^{\ast}}
\newcommand{\const}{{\rm const}}

\newcommand{\sft}{Quantum Intelligence Lab, Supremacy Future Technologies, Guangzhou 511340, China}
\newcommand{\sutd}{Science and Math Cluster and EPD, Singapore University of Technology and Design, 8 Somapah Road, 487372 Singapore}

\begin{document}

\title{A scheme for automatic differentiation of complex loss functions}
\author{Chu Guo}
\affiliation{\sft}

\author{Dario Poletti} 
\affiliation{\sutd}

\begin{abstract}
For a real function, automatic differentiation is such a standard algorithm used to efficiently compute its gradient, that it is integrated in various neural network frameworks. 
However, despite the recent advances in using complex functions in machine learning and the well-established usefulness of automatic differentiation, the support of automatic differentiation for complex functions is not as well-established and widespread as for real functions.
In this work we propose an efficient and seamless scheme to implement automatic differentiation for complex functions, which is a compatible generalization of the current scheme for real functions. This scheme can significantly simplify the implementation of neural networks which use complex numbers.         
\end{abstract}

\date{\today}
\maketitle

\address{} 

\vspace{8mm}

\section{introduction} 
Uncountable problems in mathematics, science, computer science and engineering can be formulated as finding the minimum of a function $F$. For instance, the \textit{minimum total potential energy principle} in physics states that the stationary point of a static system is the one that minimizes the potential energy~\cite{Gurtin1982}. In machine learning one often has to minimize a so-called loss function.  
Finding the minimum of a function often entails evaluating the gradients of the function itself. However, given the typical complexity of the functions to be minimized, computing its gradients can be a very difficult task. One can follow four different approaches~\cite{Baydin2017, Margossian2019}: 1) derive manually the analytic expression for the gradient, which would result in efficient calculations but which is not scalable for complicated loss functions; 2) estimate the gradient via finite-differences methods~\cite{li2005general}, which is also not scalable because it requires about $N$ evaluations of an $N-$paramenters function in order to compute the gradients while, at the same time, losing accuracy due to numerical truncation and round off errors~\cite{Jerrell1997}; 3) let the computer perform a symbolic differentiation, which returns an expression for evaluating the gradient of a function, but it does so at a large memory cost and shows limitations when used in control flows~\cite{Corliss1988}; 4) evaluation of the gradient with automatic or algorithmic differentiation (AD) which relies on considering the main function as a composite function of several \textit{elementary functions} with known gradients, which then turns the computation of the overall gradient into simple function evaluations by using the chain rule of differential calculus~\cite{Rall1996,Verma2000,Griewank2008}. 
Automatic differentiation can be further divided into two modes: a) Forward-mode AD which is suited for functions with a single input variable and multiple output variables. This can be easily implemented using the dual-number strategy~\cite{Rall1996}, i.e. any variable $v$ is stored as a tuple $\left(v, v'\right)$, where the second element is the derivative. When one evaluates a function $f(v)$, the derivative $f'(v)$ is evaluated at the same time and stored in the result; b) Reverse-mode AD which requires much more effort to implement compared to the forward-mode, but is efficient for functions with multiple input variables and a single output variable, thus ideal for loss functions whose output is a single real number. In this work we focus on reverse-mode AD while referring to it simply as AD.

While in many applications the function to be minimized is $F$: $\Rdomin^n\rightarrow\Rdomin$, i.e. from a vector of real numbers to a real scalar, more generically the function could be $F$: $\Cdomin^n\rightarrow\Rdomin$, i.e. from a vector of complex numbers to a real scalar. The use of such complex functions is very natural in quantum physics, however their use is at the center of an increasing number of investigations in the machine learning community. For instance, recurrent neural networks (RNN) generally suffer from the vanishing or exploding gradient problem \cite{Hochreiter1991, Bengio1994}, something that can be cured by the use of unitary recurrent neural networks (uRNN) which use complex unitary matrices~\cite{Arjovsky2016, Wisdom2016full, Jing2017, Trabelsi2017, Wolter2018, Maduranga2019}.    
While basic theory for the use of complex numbers for activation functions, gradients, Hessians and back-propagation \cite{Leung1991, Benvenuto1992, Georgiou1992} has been long-established, only recently the use of complex numbers has shown potential to enable easier optimization \cite{Nitta2002}, noise-robust memory mechanisms \cite{Danihelka2016}, a richer representational capacity \cite{Arjovsky2016, Wisdom2016full}, faster learning \cite{Arjovsky2016} and better generalization characteristics \cite{Hirose2012}. Complex numbers have also been used in an LSTM \cite{Hochreiter1997} architecture \cite{Danihelka2016}.

The understanding of AD for a real loss function $F$: $\Rdomin^n\rightarrow\Rdomin$ is mature enough that general-purpose AD has already been integrated into numerical frameworks, such as PyTorch~\cite{Pytorch2017}, TensorFlow~\cite{Tensorflow2016}, Chainer~\cite{Chainer2015}, Autograd~\cite{Maclaurin2016} and Zygote~\cite{Zygote2019}. However, despite the recent interest in using complex numbers, such tools are not as mature for complex functions $F$: $\Cdomin^n\rightarrow\Rdomin$. For instance, PyTorch currently does not support complex numbers, TensorFlow provides a separate interface specifically for complex numbers, and while the Zygote package is written in Julia language~\cite{Julia2017} and it has a native support for complex numbers, currently it may not, as we will explain in detail later, return the correct gradients required by the gradient-based optimizers.
While one could avoid the usage of complex numbers by treating them as tuples of two real numbers, this would result in having to manually code the gradients of all complex functions resulting in making it highly non-trivial to build a general-purpose AD package which supports complex functions.

In this work, we propose a generalized automatic differentiation scheme which provides a \textit{unified interface} for both real and complex functions. For each elementary complex function $g$, we show that one only needs to supply a slightly modified adjoint function which makes use of the Wirtinger derivatives \cite{Fichera1986, Kracht1988}, and then one could obtain the correct complex gradients of the final loss function $F$ with no additional non-automatic effort. Our approach is fully compatible with current methods in that if the same function $g$ takes real numbers as input, the generalized adjoint function will simply reduce to the standard adjoint function for real functions~\cite{Baydin2017}.

In the following we start by briefly reviewing the current implementation of automatic differentiation for real loss functions in Sec.~\ref{sec:realad}. Then in Sec.~\ref{sec:complexad}, we present our definition of generalized adjoint function which extends automatic differentiation to the complex domain. In Sec.~\ref{sec:examples}, we give some explicit examples of our generalized adjoint function for some commonly used holomorphic and non-holomorphic functions and discuss some possible applications of complex AD. We discussion our conclusions in Sec.~\ref{sec:summary}. Throughout this work the same symbol may be used for both scalars or arrays, and the exact meaning can be determined by the number of subscripts. To avoid confusion, the word ``complex'' always refers to complex numbers or functions with complex numbers involved.

\section{Review of automatic differentiation for real functions}\label{sec:realad}

\begin{figure}
\includegraphics[width=\columnwidth]{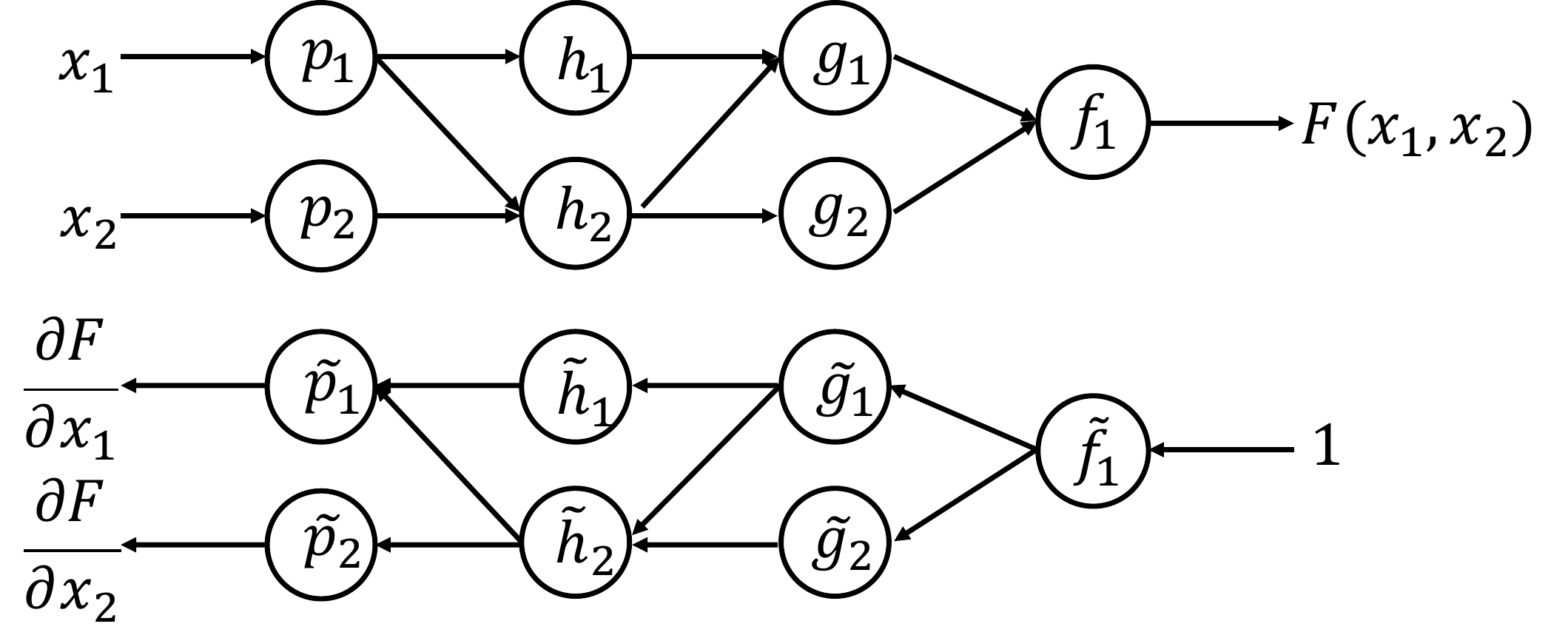}
\caption{(a) Forward evaluation of a function $F(x_1, x_2)$ of two variables $x_1$ and $x_2$. (b) Backward evaluation of the gradient of the function $F(x_1, x_2)$.}   \label{fig:fig1} 
\end{figure}

For completeness and easier understanding, we first briefly review AD for a real function $F$: $\Rdomin^n\rightarrow\Rdomin$. $F(x)$ can be viewed as a composite function built from some elementary functions for which the gradients are known. In Fig.~\ref{fig:fig1} we explicitly show an instance of a two-variable loss function. The function value $F(x)$ can be derived by sequentially evaluating each elementary function, which is often referred to as the \textit{forward process} and is shown in Fig.~\ref{fig:fig1}(a). Mathematically, we can write a generic composite function $F(x)$ as
\begin{align}\label{eq:F}
F = f \circ g \circ h \circ \dots \circ p (x).
\end{align}
We first consider the case that $x$ is a single real number and that $f$, $g$, $h$, $p$ are all scalar functions. The derivative of $F(x)$, denoted as $d F/dx$, can be computed using the well-known chain rule 
\begin{align}\label{eq:realchainrule}
\frac{d F}{d x}= \frac{d f}{d g} \frac{d g}{d h} \dots \frac{d p}{d x}. 
\end{align}
To evaluate Eq.(\ref{eq:realchainrule}) automatically, a key idea is to define an adjoint function $\adj{g}$ for each elementary function $g$ as
\begin{align}\label{eq:realadjoint}
\adj{g}(\nu)\vert_x = \nu \frac{d g}{d x}.
\end{align}
In general, the input of the adjoint function $\adj{g}$ has the same number of elements as the output of $g$, and the output of $\adj{g}$ has the same number of elements as the input of $g$. We can also see that in the definition of $\adj{g}$, the only information one needs to known from $g$ is the derivative $\frac{d g}{d x}$. With Eq.~(\ref{eq:realadjoint}), we can rewrite Eq.~(\ref{eq:realchainrule}) as 
\begin{align}\label{eq:realad}
\frac{d F}{d x} = \adj{p} \circ \dots \circ \adj{h} \circ \adj{g} \circ  \adj{f}(1).
\end{align}
By successively evaluating the adjoint function of each elementary function in the reverse order, we obtain the gradient of the loss function which is equivalent to the one computed by using the chain rule. 

In the general case, the input of $F(x)$ is an array of scalars and $f$, $g$, $h$, $p$ are vector functions instead of scalar functions, as shown in Fig.~\ref{fig:fig1}(b). As a result, the adjoint function $\adj{g}$ will be a linear mapping and Eq.~(\ref{eq:realadjoint}) can be generalized straightforwardly as 
\begin{align}\label{eq:realvectoradjoint}
\adj{g}_j(\nu)\vert_x = \sum_i \nu_i \frac{\partial g_i}{\partial x_j}, 
\end{align}
and we can seamlessly write  
\begin{align}\label{eq:realad_nabla}
\nabla F = \adj{p} \circ \dots \circ \adj{h} \circ \adj{g} \circ  \adj{f}(1).
\end{align}
Eq.~(\ref{eq:realad_nabla}) is also referred to as the \textit{backward process}, in juxtaposition with the forward process. The advantage of this approach is that one can predefine the adjoint functions for a finite set of elementary functions and then let the program ``automatically'' compute the gradient of any function built from those elementary functions with Eqs.~(\ref{eq:realad}, \ref{eq:realad_nabla}). 

We stress here that the performance of AD will be strongly affected by the way that the elementary functions are chosen. Using too many elementary functions would be more inefficient and require more memory since a larger amount of intermediate variables have to be stored. Using a much smaller number of elementary functions would make the algorithm less ``automatic''. Therefore, in real applications one needs to balance between performance and coding simplicity.

\section{Automatic Differentiation for complex functions}\label{sec:complexad}
In the following we generalize Eqs.(\ref{eq:realadjoint}, \ref{eq:realvectoradjoint}) to the case of complex functions. We first consider that the loss function is a function of a single complex number $z$ and its complex conjugate $\conj{z}$, which can be written as $F(z, \conj{z})$, and we will generalize our results to arbitrary inputs in a second moment. Since $F$ is a mapping from a complex number to a real number, it is in general a non-holomorphic function, namely $\frac{\partial F}{\partial \conj{z}} \neq  0$.  
We thus use the standard complex derivative of a complex function $F$ as 
\begin{align}\label{eq:complexderivative}
dF = \frac{\partial F}{\partial z} dz + \frac{\partial F}{\partial \conj{z}} d\conj{z},
\end{align}
where we have used the Wirtinger derivatives $\frac{\partial F}{\partial z}$ and $\frac{\partial F}{\partial \conj{z}}$ defined as
\begin{align} 
\frac{\partial F}{\partial z} &= \frac{\partial F(z, \conj{z})}{\partial z}|_{\conj{z}=\const} = \frac{1}{2}\left(\frac{\partial F}{\partial x} - \im \frac{\partial F}{\partial y} \right); \label{eq:formala} \\
\frac{\partial F}{\partial \conj{z}} &= \frac{\partial F(z, \conj{z})}{\partial \conj{z}}|_{z=\const} = \frac{1}{2}\left(\frac{\partial F}{\partial x} + \im \frac{\partial F}{\partial y} \right). \label{eq:formalb}
\end{align}
Note that the first equality in Eqs.(\ref{eq:formala},\ref{eq:formalb}) is a {\it formal} definition made rigorous by the second identity.

The chain rule in Eq.~(\ref{eq:realchainrule}) can be straightforwardly generalized to the case of a complex loss function $F(z, \conj{z})$ using the Wirtinger derivatives
\begin{align}\label{eq:complexchainrule}
\left[\begin{array}{cc}
\frac{\partial F}{\partial z} & \frac{\partial F}{\partial \conj{z}}
\end{array}\right] =& \left[\begin{array}{cc}
\frac{\partial f}{\partial g} & \frac{\partial f}{\partial \conj{g}}
\end{array}\right] \times  \left[\begin{array}{cc}
\frac{\partial g}{\partial h} & \frac{\partial g}{\partial \conj{h}} \\
\frac{\partial \conj{g}}{\partial h} & \frac{\partial \conj{g}}{\partial \conj{h}} 
\end{array}\right] \times \nonumber \\ 
 & \dots \times\left[\begin{array}{cc}
\frac{\partial p}{\partial z} & \frac{\partial p}{\partial \conj{z}} \\
\frac{\partial \conj{p}}{\partial z} & \frac{\partial \conj{p}}{\partial \conj{z}} 
\end{array}\right].
\end{align}
Similar to the real case in Sec.~\ref{sec:realad}, we can define an adjoint function $\adj{g}$ for a complex function $g$ as
\begin{align}\label{eq:genericcomplexadjoint}
\adj{g}(\left[\begin{array}{cc}
\nu & \conj{\nu}
\end{array}\right])\vert_z &= \left[\begin{array}{cc}
\nu & \conj{\nu}
\end{array}\right] \times \left[\begin{array}{cc}
\frac{\partial g}{\partial z} & \frac{\partial g}{\partial \conj{z}} \\
\frac{\partial \conj{g}}{\partial z} & \frac{\partial \conj{g}}{\partial \conj{z}}
\end{array}\right] \nonumber \\
&= \left[\begin{array}{cc}
\nu \frac{\partial g}{\partial z} + \conj{\nu} \frac{\partial \conj{g}}{\partial z} & \nu \frac{\partial g}{\partial \conj{z}} + \conj{\nu} \frac{\partial \conj{g}}{\partial \conj{z}}
\end{array}\right],
\end{align}
which takes a pair of complex conjugates as input instead of a single real number in Eq.~(\ref{eq:realadjoint}). With Eq.~(\ref{eq:genericcomplexadjoint}) we can verify that 
\begin{align}\label{eq:genericcomplexad}
\left[\begin{array}{cc}
2\frac{\partial F}{\partial z} & 2\frac{\partial F}{\partial \conj{z}}
\end{array}\right] = \adj{p} \circ \dots \circ \adj{h} \circ \adj{g} \circ  \adj{f}(\left[\begin{array}{cc}
1 & 1
\end{array}\right]),
\end{align}
where we have used the fact that the output of the last function $f$ is real, namely $f=\conj{f}$.  
Eq.~(\ref{eq:genericcomplexad}) can be used for any complex loss function $F(z, \conj{z})$, however, the input and output of the complex adjoint function in Eq.~(\ref{eq:genericcomplexadjoint}) are tuples of two elements in comparison with the real case in Eq.~(\ref{eq:realadjoint}). So, at this point, it seems that the approach with complex functions may not seemleasly include the case for real functions. 

However here we should do a step back. It is important to notice that Eq.~(\ref{eq:complexderivative}) does not directly tell us the correct gradient required by the gradient based optimization algorithms as in the real case~\cite{Kreutz2009}. This is because in Eqs.~(\ref{eq:formala}, \ref{eq:formalb}) $z$ and $\conj{z}$ are not independent variables, namely one cannot change $z$ while keeping $\conj{z}$ constant. Essentially speaking, $F(z, \conj{z})$ is just a function of $z$ since $\conj{z}$ is dependent on $z$. To obtain the correct gradient required by gradient-based optimizers, one should think of $z=x+\im y$ as a tuple of two real numbers, $z = (x, y)$, because $x$ and $y$ are independent variables. As a result, the loss function becomes a function of $x$ and $y$ which can be written as $F(x, y)$, and the gradient of $F(x, y)$ is a tuple of partial derivatives $\left(\frac{\partial F}{\partial x}, \frac{\partial F}{\partial y} \right)$. Transforming back into the complex domain, the correct gradient required by gradient-based methods is thus $2\frac{\partial F}{\partial \conj{z}}$ (for a proof one can also refer to~\cite{Boeddeker2017}). To demonstrate this in more detail, we take the gradient descent algorithm as an example. We assume a learning rate of $\lambda$ where $\lambda$ is a small positive real number. Then, in one iteration of the gradient descent algorithm, the real variables $(x, y)$ would be updated as $(x - \lambda \frac{\partial F}{\partial x}, y - \lambda \frac{\partial F}{\partial y})$. The resulting value of the loss function will decrease by 
\begin{align}
&F(x - \lambda \frac{\partial F}{\partial x}, y - \lambda \frac{\partial F}{\partial y} ) - F(x, y) \nonumber \\  
= &-\lambda\left[\left(\frac{\partial F}{\partial x}\right)^2 + \left(\frac{\partial F}{\partial y}\right)^2\right] + O(\lambda^2).
\end{align}
If $F$ is treated as a function of $z$ and $\conj{z}$ instead, then it is easy to verify that if $z$ is updated as $z -2\lambda \frac{\partial F}{\partial \conj{z}}$, one would have
\begin{align}
&F(z - 2\lambda \frac{\partial F}{\partial \conj{z}}, \conj{z} - 2\lambda \frac{\partial F}{\partial z} ) - F(z, \conj{z}) \nonumber \\ 
= &-4\lambda \frac{\partial F}{\partial z} \frac{\partial F}{\partial \conj{z}} + O(\lambda^2) \nonumber \\ 
=& -\lambda\left[\left(\frac{\partial F}{\partial x}\right)^2 + \left(\frac{\partial F}{\partial y}\right)^2\right] + O(\lambda^2) ,
\end{align}
where we have used Eqs.(\ref{eq:formala}, \ref{eq:formalb}) and the fact that $\conj{z}$ will be updated to $\conj{z} -2\lambda \frac{\partial F}{\partial z}$ accordingly since it is dependent on $z$.

We can thus define a simplified complex adjoint function which only contains half of the pairs or, in other words, we can simplify Eq.~(\ref{eq:genericcomplexadjoint}) to be 
\begin{align}\label{eq:complexadjoint}
\adj{g}(\conj{\nu})\vert_z = \nu \frac{\partial g}{\partial \conj{z}} + \conj{\nu} \frac{\partial \conj{g}}{\partial \conj{z}},
\end{align}
Note that if $g$ is a real function, namely $\frac{\partial g}{\partial z} = \frac{\partial g}{\partial \conj{z}} = \frac{1}{2}\frac{\partial g}{\partial x}$, then Eq.~(\ref{eq:complexadjoint}) reduces to $\adj{g}(\nu) = \nu \frac{\partial g}{\partial x}$, which is exactly Eq.~(\ref{eq:realadjoint}). Therefore, the definition of complex adjoint function in Eq.~(\ref{eq:complexadjoint}) is fully compatible with the real case. This implies that one can simply substitute the definition of adjoint function in Eq.~(\ref{eq:realadjoint}) by Eq.~(\ref{eq:complexadjoint}) and AD will work perfectly for both real and complex functions with negligible computational overhead. Based on Eq.~(\ref{eq:complexadjoint}), Eq.~(\ref{eq:genericcomplexad}) can be simplified as
\begin{align}\label{eq:complexad}
2\frac{\partial F}{\partial \conj{z}} = \adj{p} \circ \dots \circ \adj{h} \circ \adj{g} \circ  \adj{f}(1),
\end{align}
which has the same form as the real case in Eq.~(\ref{eq:realad}). 

For a non-scalar complex function $g$, Eq.~(\ref{eq:complexadjoint}) can be straightforwardly generalized to 
\begin{align}\label{eq:complexvectoradjoint}
\adj{g}_j(\conj{\nu})\vert_z = \sum_i\left(\nu_i \frac{\partial g_i}{\partial \conj{z}_j} + \conj{\nu}_i \frac{\partial \conj{g}_i}{\partial \conj{z}_j}\right).
\end{align}
Eqs.(\ref{eq:complexadjoint}, \ref{eq:complexad}, \ref{eq:complexvectoradjoint}) constitute the main results of this work.

\section{Examples and Applications}\label{sec:examples}

\subsection{Examples of complex gradients and some useful properties}\label{sec:list}             

In~\cref{tab:examples} we list some explicit examples as direct applications of Eqs.(\ref{eq:complexadjoint}, \ref{eq:complexvectoradjoint}). Table \ref{tab:examples} shows that our definition of adjoint functions is applicable for both real and complex inputs. 

\begin{table}[!htb]
\centering
\caption{{\bf Adjoint functions of some common functions.} The column denoted by ``Function'' is the name of the function. The column denoted by ``Forward'' is the definition of the function, and the column denoted by ``Backward'' is the adjoint function of the original function. Symbols such as $z$ or $w$ without subscript indicate a (complex) scalar. $z_i$ with a single subscript indicates a vector and $z_{ij}$ a matrix. Einstein summation notation is used in this table. $N$ in the Fourier and inverse Fourier transformations inidcates the length of the input vector.\\}
\label{tab:examples}
\begin{tabular}{|l|l|r|}
\hline
{\bf Function } & {\bf Forward }& {\bf Backward} \\
 \hline
sine & $z \rightarrow \sin(z)$ & $\conj{\nu}\rightarrow \conj{\nu}\cos(\conj{z})$ \\
scalar exponential & $z \rightarrow e^z$ & $\conj{\nu}\rightarrow \conj{\nu}e^{\conj{z}}$ \\
scalar logarithm & $z \rightarrow \log(z)$ & $\conj{\nu}\rightarrow \frac{\conj{\nu}}{\conj{z}}$ \\
scalar addition & $\left(z, w\right)\rightarrow z+w$ & $\conj{\nu}\rightarrow\left(\conj{\nu}, \conj{\nu}\right)$ \\
scalars multiplic. & $\left(z, w\right)\rightarrow z w$ & $\conj{\nu}\rightarrow\left(\conj{\nu}\conj{w}, \conj{\nu}\conj{z}\right)$ \\
scalars division & $\left(z, w\right)\rightarrow \frac{z}{w}$ & $\conj{\nu}\rightarrow \left(\frac{\conj{\nu}}{\conj{w}}, -\frac{\conj{\nu} \conj{z}}{{\conj{w}}^2}  \right)$ \\
real part & $z\rightarrow \real(z)$ & $\conj{\nu}\rightarrow \real(\conj{\nu})$ \\
imaginary part & $z\rightarrow \imag(z)$ & $\conj{\nu}\rightarrow \im\real(\conj{\nu})$ \\
absolute value & $z\rightarrow \vert z\vert$ & $\conj{\nu}\rightarrow \real(\nu)\frac{z}{\vert z\vert}$ \\
inner product & $\left(z_i, w_j\right) \rightarrow \conj{z}_i w_i$ & $\conj{\nu}\rightarrow \left(\nu w_i, \conj{\nu} z_j  \right)$ \\
outer product & $\left(z_i, w_j\right) \rightarrow z_i w_j$ & $\conj{\nu}_{ij}\rightarrow \left(\conj{\nu}_{ij} \conj{w}_j, \conj{z}_i\conj{\nu}_{ij}  \right)$ \\
matrix multiplic. & $\left(z_{ij}, w_{jk}\right) \rightarrow z_{ij}w_{jk}$ & $\conj{\nu}_{ik}\rightarrow \left(\conj{\nu}_{ik}\conj{w}_{jk}, \conj{z}_{ij} \conj{\nu}_{ik} \right)$ \\
Fourier & $z_n \rightarrow e^{-\frac{2\pi \im}{N}kn}z_n$ & $\conj{\nu}_{k}\rightarrow \conj{\nu}_ke^{\frac{2\pi \im}{N}kn}$ \\
inverse Fourier & $z_k \rightarrow \frac{1}{N}e^{\frac{2\pi \im}{N}kn}z_k$ & $\conj{\nu}_{n}\rightarrow \frac{1}{N}\conj{\nu}_ne^{-\frac{2\pi \im}{N}kn}$ \\
\hline
\end{tabular}
\end{table}

We also point out that while evaluating Eq.~(\ref{eq:complexadjoint}) in general requires to compute both $\frac{\partial g}{\partial z}$ and $\frac{\partial g}{\partial \conj{z}}$, there are several special cases in which only $\frac{\partial g}{\partial z}$ or $\frac{\partial g}{\partial \conj{z}}$ need to be evaluated to derive Eq.~(\ref{eq:complexadjoint}). First, if $g$ is a holomorphic function, then Eq.~(\ref{eq:complexadjoint}) reduces to
\begin{align}\label{eq:cadjs}
\adj{g}(\conj{\nu})\vert_z = \conj{\nu} \frac{\partial \conj{g}}{\partial \conj{z}} = \conj{\nu} \conj{\left(\frac{\partial g}{\partial z}\right)},
\end{align}
and if $g$ is an anti-holomorphic function we have 
\begin{align}\label{eq:cadjs}
\adj{g}(\conj{\nu})\vert_z = \nu \frac{\partial g}{\partial \conj{z}}.
\end{align}
Second, if the input of $g$ is real, that is $g(x) = g\left(\frac{1}{2}(z+\conj{z})\right)$, then 
$\conj{\left(\frac{\partial g}{\partial z}\right)} = \conj{\left(\frac{\partial g}{\partial \conj{z}}\right)}=\frac{1}{2} \frac{\partial g}{\partial x} $ and we have
\begin{align}\label{eq:cadjs}
\adj{g}(\conj{\nu})\vert_z = 2\real\left(\nu \frac{\partial g}{\partial \conj{z}}\right),
\end{align}
where $\real(z)$ means to take the real part of a complex number $z$. Lastly, if the output of $g$ is real, i.e. $g=\conj{g}$, then $\frac{\partial \conj{g}}{\partial \conj{z}} = \frac{\partial g}{\partial \conj{z}}$ and we have
\begin{align}\label{eq:cadjs}
\adj{g}(\conj{\nu})\vert_z = 2\real(\nu) \frac{\partial g}{\partial \conj{z}}.
\end{align}

\subsection{Comparison to using a tuple of real and imaginary part of a complex number}  

An approach which is currently used in the study of neural networks is to treat a complex number by splitting it into a tuple of two real numbers, and then redefine the corresponding functions in terms of the resulting tuple accordingly. To show how this approach compares to the one we propose here, we consider a simple example in which we take a complex function of two complex numbers $z$ and $w$ 
\begin{align}\label{eq:simpleexample}
g(z, w) = z w,
\end{align}
where $z=a_z+\im b_z$, $w=a_w+\im b_w$ and $a_z$, $b_z$, $a_w$ and $b_w$ are real. From Eq.~(\ref{eq:complexadjoint}), the adjoint function of $g$ is simply given by 
\begin{align}
\adj{g}(\conj{\nu})\vert_{z, w} = \left(\conj{\nu}\conj{w}, \conj{\nu}\conj{z} \right),
\end{align}
where $\nu = u + \im v$ is another complex variable. If, instead, $z$ and $w$ are each treated as tuples and one use Eq.~(\ref{eq:realadjoint}) instead, then the adjoint function of $g$ should be defined as
\begin{align}
\adj{g}\left(\left(u, v\right)\right)\vert_{(a_z,b_z), (a_w, b_w)} =\left( \left(ua_w + vb_w, va_w-ub_w \right), \right. \nonumber \\ \left. \left(ua_z + vb_z, va_w-ub_w  \right) \right), 
\end{align} 
which is already fairly cumbersome to compute even for this simple case.\\ 
A different approach is to view $g$ as a composite function of ``more'' elementary functions as
\begin{align}  
g(z, w) = \left(\real(z)\real(w) - \imag(z)\imag(w), \right. \nonumber \\ \left. \real(z)\imag(w) + \imag(z)\real(w) \right), \label{eq:reim} 
\end{align} 
where, in fact, one needs to use the functions $\real(z)$ and $\imag(z)$ where the latter gives the imaginary part of a complex number $z$. From Eq.(\ref{eq:reim}) one can define the adjoint functions of $\real(\cdot)$ and $\imag(\cdot)$ as     
\begin{align}
&\adj{\real}(\nu)\vert_z = \left(\nu, 0\right) \\
&\adj{\imag}(\nu)\vert_z = \left(0, \nu\right),
\end{align}
where $\nu$ is a real number. Then as long as one has defined the adjoint functions of real arithmetics, the gradient of $g$ would be automatically derived with AD for real functions. This approach can, in principle, allow AD for real functions to work with complex functions with less non-automatic work. However, as this simple example already shows, this approach results in deeper nested functions in the backward process, which would usually consume more memory and reduce the efficiency of the computation. We highlight here that the Zygote~\cite{Zygote2019} package internally treats complex numbers as tuples and automatically derives complex gradients for them, however not all the gradients defined in Zygote comply with Eq.(\ref{eq:complexvectoradjoint}). For instance, the adjoint function of the vector inner dot product function is defined as $\conj{\nu}\rightarrow\left(\conj{\nu} w_i, \conj{\nu} z_j \right)$ in Zygote, which differs from the expression in Table~\ref{tab:examples}. Hence, it can result in incorrect gradients for complex loss functions.

\subsection{Unitary Recurrent Neural Networks}\label{sec:RNN}

As mentioned in the introduction, complex numbers become very convenient when dealing with unitary Recurrent Neural Networks, which provide a potent way to overcome the vanishing or exploding gradient problem.  
In general, the input for a RNN layer is a sequence of data. The action of the RNN on one of the input sequence $x_t$ (where $t$ labels the position in the sequence) can be written as
\begin{align}
h_t &= \sigma(Wh_{t-1} + Vx_t);\\
y_t &= Uh_t + c,
\end{align}
where $W$, $V$ and $U$ are matrices, $h_t$ is the $t$-th hidden state, $y_t$ is the $t$-th data of the output sequence, $c$ is the bias and $\sigma$ is the (nonlinear) activation function. 

In~\cite{Arjovsky2016} the authors parametrized the $W$ matrix as a unitary matrix, and they chose a particular parametrization of such matrix which is given by 
\begin{align}\label{eq:Wmat}
W = D_3 R_2 \mathcal{F}^{-1} D_2 \Pi R_1 \mathcal{F} D_1,
\end{align}
where $D_{\sigma}$ are diagonal matrices with diagonal elements $D_{\sigma,j,j}=e^{\im\omega_j}$, $R_{\sigma}$ are reflection matrices defined as 
\begin{align}
R_{\sigma} = I - 2 \frac{v_{\sigma}\conj{v}_{\sigma}}{|| v_{\sigma}||^2},
\end{align}
with $v_{\sigma}$ a complex vector and ${\sigma}=1,2,3$. 
$\Pi$ is a fixed random permutation matrix and $\mathcal{F}$, $\mathcal{F}^{-1}$ are the Fourier and inverse Fourier transformations. Therefore $W$ is parameterized by three real vector plus two complex vectors. In~\cite{Arjovsky2016}, the authors splitted each complex vector into two real vectors and redefined the multiplication in Eq.~(\ref{eq:Wmat}) accordingly. However, with our scheme for complex AD, the gradients of $W$ could be simply derived in the same way as the real case without special handling of complex functions, as long as we define the adjoint functions for some elementary complex functions. 

Since the parametrization in Eq.(\ref{eq:Wmat}) is not all encompassing, in~\cite{Wisdom2016full}, the authors proposed to directly parameterize $W$ as a full unitary matrix, and proposed a different way to update $W$ given by 
\begin{align}
W \leftarrow \left(I + \frac{\lambda}{2}A\right)^{-1}\left(I - \frac{\lambda}{2}A\right) W,
\end{align}
where $\lambda$ is the learning rate, $A_{ik} = \frac{\partial F}{\partial \conj{W}_{ij}}W_{ik} - \conj{W}_{ij} \frac{\partial F}{\partial W_{ij}}$ with $F(W)$ the loss function. To compute $\frac{\partial F}{\partial W_{ij}}$, $W$ was splitted into a tuple of two real matrices as $\left(\real(W), \imag(W)\right)$ and then partial derivatives against the real and imaginary parts were computed using real AD respectively. 
For this to work, one has to redefine all the linear algebra functions along the way to act on real tuples. When the loss function $F(W)$ becomes more complicated or there are more complex units in the deep neural network, this redefining process would cause a sizeable amount of non-automated work. With the complex AD discussed here, computing $\frac{\partial F}{\partial W_{ij}}$ would just be as easy as for the real case. 

\section{Conclusion}\label{sec:summary}
In summary, we have presented a scheme which generalizes current automatic differentiation to work in the complex domain. For functions with complex input, our scheme relies on an adjoint function which has a similar form as the current definition for real input, but is able to derive the correct gradients of a generic loss function required by gradient-based optimizers. 
While the AD scheme we presented will not necessarily improve the performance of existing codes, it should significantly simplify the coding part, thus possibly resulting in an increased use of, for instance, unitary recurrent neural networks or, more generally, the seamless development of AD applications containing complex functions.    
This scheme could be integrated into mainstream automatic differentiation frameworks which can use complex numbers (e.g. running on Julia). 
For the interested readers, we have uploaded examples related to Sec.~\ref{sec:examples} of AD codes which uses our scheme for complex numbers in ${\rm https://github.com/guochu/complexAD}$. Moreover, the open source Julia package VQC (${\rm https://github.com/guochu/VQC.jl}$), which performs auto differentiation for quantum circuits with parametric quantum gate operations, is based on the scheme of this work.

\begin{acknowledgments} 
We thank S. Lin for fruitful discussion. C. G. acknowledges support from National Natural Science Foundation of China under Grants No. 11805279. D.P. acknowledges support from Ministry of Education of Singapore AcRF MOE Tier-II (project MOE2018-T2-2-142).   
\end{acknowledgments}

\bibliographystyle{apsrev4-1}
\bibliography{refs}

\end{document}